\pdfoutput=1
\documentclass[11pt]{article}

\usepackage[review]{acl}

\usepackage{times}
\usepackage{latexsym}
\usepackage{adjustbox}
\usepackage{titlesec}
\usepackage[T1]{fontenc}
\usepackage[utf8]{inputenc}

\usepackage{microtype}
\definecolor{movie_color}{RGB}{223, 84, 65}
\newcommand{\movie}[1]{\textcolor{movie_color}{\textbf{#1}}}
\usepackage{mdframed}
\title{Multi-Task End-to-End Training Improves Conversational Recommendation}
\author{Naveen Ram, Dima Kuzmin, Ellie Ka In Chio \\  \bf{Moustafa Farid Alzantot, Santiago Ontanon, Ambarish Jash, \and Judith Yue Li}\\
      Google Research \\ \texttt{[naveenram,dimakuzmin,echio,mfarid,santiontanon,ajash,judithyueli]@google.com}}
        
\begin{document}
\nolinenumbers
{\makeatletter\acl@finalcopytrue
  \maketitle
}
\begin{abstract}

In this paper, we analyze the performance of a multitask end-to-end transformer model on the task of conversational recommendations, which aim to provide recommendations based on a user’s explicit preferences expressed in dialogue. While previous works in this area adopt complex multi-component approaches where the dialogue management and entity recommendation tasks are handled by separate components, we show that a unified transformer model, based on the T5 text-to-text transformer model, can perform competitively in both recommending relevant items and generating conversation dialogue. We fine-tune our model on the ReDIAL conversational movie recommendation dataset, and create additional training tasks derived from MovieLens (such as the prediction of movie attributes and related movies based on an input movie), in a multitask learning setting. Using a series of probe studies, we demonstrate that the learned knowledge in the additional tasks is transferred to the conversational setting, where each task leads to a $9\% - 52\%$ increase in its related probe score.

\end{abstract}

\section{Introduction}
The modern recommendation systems found in commercial applications are largely based on implicit preferences, such as a user's history of web page clicks, item purchases, or media streams, with the record of these actions used to retrieve relevant recommendations \cite{rendle2012bpr}. This approach often works, but in the case where a user might not have an extensive history, or might desire a recommendation which doesn't match their usual niche, we might want a system which can take advantage of explicit preferences. With the growing success of deep learning language models, it has become possible to design conversational recommendation models which can communicate with a user directly while retrieving custom recommendations based on the user's explicit preferences. 

Most previous work on conversational recommender systems adopts a multi-component approach \citep{gao2021advances}. These models often are implemented using a recommendation component, which analyzes the mentioned entities in order to predict a related item, and a dialogue component, which analyzes the input phrases and generates a conversational response \cite{dietmar2020survey}. Multi-component approaches are appealing because they can be built directly from standard models in the dialogue and recommendation fields. However, the knowledge learned by each component is not immediately available to the other components (i.e., the item recommendation model does not benefit directly from conversation state, and vice versa), preventing these approaches from taking advantage of the data to its fullest extent. Ideally, a conversational recommendation model should be able to both use descriptive language in the dialogue to retrieved relevant items and generate engaging dialogue about the items simultaneously. To address this problem, in this paper we investigate whether an end-to-end approach to conversational recommendations using a single component model can improve dialogue and recommendation generation by allowing the model to fully utilize the conversation features for both tasks.

This paper strives to show the feasibility of a unified model for conversational recommendations by leveraging a single large transformer model to generate both relevant recommendations and natural dialogue and evaluating the benefits of a fully unified dialogue and recommendation module. To determine whether single end-to-end model match or outperform multi-component approaches, we train our model on several standard datasets in the domain of movie recommendations and compare our results to previous work. To measure the benefit of generating dialogue and recommendations in the same model, we follow a common procedure in related work~\cite{penha2020what} and  design a series of {\em probes} to assess how the model leverages different types of information to generate dialogue and recommendations. One potential problem of a single component system is the reliance on a large dataset of sample dialogues containing both recommendation and language information. To bypass the need for a single large dialogue dataset, we finetune the pretrained T5 model on a relatively small dataset of dialogues, and incorporate movie relationship, attribute, and description information from additional datasets using a multitask setup. 

The main contributions of this paper are:

\begin{itemize}
    \item A fully end-to-end approach to conversational recommendation that uses a unified model for both dialogue and item recommendation.
    \item Conducting a series of probe studies that shows how conversational recommendation tasks benefits from knowledge learned by the model via multi-task training on a number of separate, small datasets.
\end{itemize}

The remainder of this paper is structured as follows. First, we briefly present some related work on conversational recommender systems, transformer models and probes studies. After that, we describe our T5-based approach, and the datasets, tasks and training procedure used to train our model. We then describe our experimental methodology, and a series of probe studies showing how dialogue and recommendation mutually improved by sharing a common model.

\subsection{Related Work}

The section presents a brief background on conversational recommendations, multitask transformer models, and the evaluation of conversational recommendation models through probe studies. 

A recent survey published by \citet{gao2021advances} highlights the range of strategies used to address the different challenges faced by a conversational recommendation system (CRS): \textit{Question-based User Preference Elicitation}, performed by models which use natural dialogue questions to build an accurate user representation \cite{zou2020toward}, \textit{Multi-turn Conversational Recommendation Strategies}, which use various dialogue representations to keep track of user preferences over a long form dialogue \cite{li2021seamlessly}, and \textit{Natural Language Understanding and Generation}, which often relies on large pretrained language models to translate recommendations into natural text \cite{wang2021finetuning}. While these challenges have been approached using a variety of multi-component models, our model aims to demonstrate that a single-component transformer model can perform the task of conversational recommendations, and even benefit from cross-task transfer due to its unified design. A different approach to unified conversational recommendations, by \citet{deng2021unified}, succeeds in building a single-component system based on a graph-based Markov Decision Process which switches between predefined question-asking, recommendation, and conversation patterns in order to lead a user to a recommendation in a multi-turn dialogue. This approach, however, is fixed to a rigid flow of conversation patterns and does not contain natural language understanding or generation components necessary to create or understand free-form dialogues or unstructured conversations.

Dialogue generation has historically been approached in many ways, with recent efforts focusing on RNNs models (like LSTMs~\cite{hochreiter1997long}), and transformers~\citep{vaswani2017attention}. Recommendation systems typically perform collaborative filtering on a set of user-item associations using a range of models such as matrix factorization systems or autoencoders \citep{ricci2011recommender}. \citet{li2018towards} proposed an approach to combining these two areas into a functional conversational recommendation model, using an autoencoder recommender in conjunction with a GRU based hierarchical encoder-decoder system to generate the dialogue. There is some interplay between components, with mentioned movies and sentiments being fed into the autoencoder in order to retrieve a relevant recommendation based on a user's liked and disliked movies, but the generation of dialogues and recommendations are still largely separate. \citet{chen2019towards} took this approach one step further, creating a conversational recommendation system which would use mentioned entities in the dialogue to conduct a knowledge graph search of related items and add a vocabulary bias based on the user representation back into the transformer-based dialogue generation module. Although this model demonstrates the potential for transfer between dialogue and recommendation tasks, it requires a complex structure where incomplete representations of both the dialogue and recommendation features are passed to separate components and then joined with a switching network. In this paper we attempt to fully leverage this cross-task transfer without the need for separate components.

In recent years, many studies have demonstrated the effectiveness of large, pre-trained transformer-based language models on a range of natural language generation tasks. The architecture, which makes use of self attention blocks in order to model language, was proposed by \citet{vaswani2017attention} and achieved state-of-the-art performance on a variety of benchmarks. When pretrained on a large corpus of text, transformer models such as BERT \cite{devlin2018bert}, GPT-3 \cite{brown2020gpt3}, and UniT \cite{hu2021unit} have shown the ability to handle multiple language-based tasks with minimal finetuning. The T5 model, introduced by \cite{raffel2019exploring}, has demonstrated a distinct ability to incorporate different types of knowledge from multiple sources and handle several disparate tasks in the text-to-text format. 

In regards to evaluating these large transformer models on the task of conversational recommendations, one effective approach proposed by \citet{penha2020what} is to use {\em probe studies} to measure the model's ability to score the likelihood of certain entities when conditioned on a set of generated inputs. \citeauthor{penha2020what} evaluate BERT's performance on conversational recommendation tasks by using BERT's prediction, similarity, and next sentence prediction function to score the model's ability to associate a book, movie, or song with a related item or attribute. Although \citeauthor{penha2020what} evaluate BERT's knowledge on conversational recommendations, it is important to note that in their study BERT is not acting as a full conversational recommendation system on its own and cannot be considered an example of an end-to-end CRS. It is only being used to rank probes against each other, and not to generate dialogues and recommendation or manage a conversation with a user.

\section{Our Approach}

The main idea of our approach is to formulate the conversational recommendation task as an instance of the text-to-text problem. We finetune a pre-trained transformer model on the movie recommendation dialogues contained in the ReDial dataset, and improve the model's ability to utilize movie attributes and descriptive details within the dialogues through the introduction of additional training tasks in a multi-task learning setting. In this section we present a background on the T5 transformer model, a summary of each of the training datasets, and an overview of the finetuning parameters we used. 

\subsection{T5 Model}

T5 is a large, publicly available, encoder-decoder transformer based model created by \citet{raffel2019exploring}. The model was trained and structured with the intent to support as many different use cases as possible using a text-to-text format. In the context of recommendation systems, T5 and related models are attractive because they perform well on natural language understanding and generation tasks and has demonstrated the ability to train on multiple disparate types of text data within one model. 

\subsection{ReDial Dialogue Task}

\begin{table*}[tb!]\centering 
\resizebox{\textwidth}{!}{
\begin{tabular}{p{0.1\textwidth}p{0.2\textwidth}p{0.25\textwidth}p{0.18\textwidth}p{0.22\textwidth}} 
& \bf{ReDial Dialogues} & \bf{MovieLens Sequences} & \bf{MovieLens Tags} & \bf{MovieLens Reviews} \\ \hline \hline
\textbf{Training Inputs}  
& [User] I'm in the mood to watch a romantic comedy. What do you suggest? [Assistant] \movie{@ 50 First Dates (2004) @}
& \movie{@ The Incredibles (2004) @ Harry Potter and the Chamber of Secrets (2002)   The Hunger Games Mockingjay - Part 1 (2014) @} 
& drama, based on a book, adapted from:book &  Review for \movie{@ Alice in Wonderland (1951) @}:\\ \hline
\bf{Training Targets}  & Yes she is good. Do you like \movie{@ The Wedding Singer (1998) @} & \movie{Underworld: Awakening (2012)} & \movie{The Book Thief (2013)} & Perhaps because its surrealism matched the hippy culture of psychedelia, \movie{Alice in Wonderland (1951)} \textsuperscript{1} ... \\ \hline
\bf{Knowledge}  & Dialogue  & Recommendation & Attributes & Description

\end{tabular}
}

\caption{Comparison of the primary training task (ReDial Dialogues) and the three auxilary training tasks designed to increase recommendation, attribute, and description knowledge.}
\label{fig:trainingtasks} 
\end{table*}

The ReDial (Recommendation Dialogues) dataset is an annotated set of 11248 dialogues collected through Amazon Mechanical Turk \cite{li2018towards}. Each dialogue contains the movies and messages sent between two parties acting as either a "recommender" or a "recommendation seeker". Although this dataset is relatively small, and doesn't necessarily capture as much movie relationship and attribute data as other recommendation-focused datasets, we have found that it provides enough examples for the T5 to learn the style and structure of conversational recommendations.

For each conversation in the dataset we create a training example corresponding to each response from the human recommender. The model inputs contain the conversation up to a certain recommender utterance, with the outputs containing the next utterance from the recommender party. Using this format the T5 model can learn to parse relevant movie, attribute, and dialogue details from the previous messages in the conversation and formulate an appropriate response. We use the T5's standard vocabulary, so movie titles are processed by the word, the same as any other piece of the input. To help the model learn these titles, \texttt{@} signs are used to separate movie titles from the rest of the dialogues.

Each message in the ReDial conversation is preceded by either a \texttt{[USER]} or a \texttt{[ASSISTANT]} tag to indicate its source. The redial conversation shown in appendix \ref{sec:appendix} has been processed into multiple training examples corresponding to each response by the recommender. Table \ref{fig:trainingtasks} shows a sample training example from this process. 

% \footnotetext{\textsuperscript{1} }

\subsection{MovieLens Sequences Task}

The MovieLens 25m dataset is a collection of 25 million ratings and one million tag associations often used to quantify movie relationships and attributes \citep{10.1145/2827872}. We utilize this data for multiple tasks, as it can be used to quantify different types of movie information. The first additional training task is to recommend a movie given a sequence of 1-9 related movies. This task incorporates movie relationship information in order to increase the quality of recommendations in the ReDial Dialogues task, and is referred to as the ML Sequences task.

In order to use the user ratings contained in the MovieLens 25m dataset to generate movie associations, we create sequences of movies wherever there are 10 movies rated higher than 4.0 / 5.0 by the same user. From these sequences we create examples for each $n$ where $1<n<10$ by mapping the first $n$ movies as the inputs and the movie in position $(n+1)$ as the target. An example of this format is shown in Table \ref{fig:trainingtasks}.

\subsection{MovieLens Tags Task}
The MovieLens 25m dataset contains a tag genome which scores each movie's relevance across a set of 1,129 tags \cite{10.1145/2362394.2362395}. These tags are movie attributes or descriptive words which often correspond to genres ("horror", "action", "mystery"), plot elements ("alien invasion", "character study", "father daughter relationship"), opinion ("excellent script", "boring", "over the top"), or general information ("oscar (best actor)", "based on a book", "stanley kubrick"). For each movie, we add each tag with a relevance score over 0.8 to the movies tag list. From these tag lists we randomly sample examples containing 1-5 tags as the input and the related movie as a the target. This mapping allows the model to associate movie attribute information and better recommend movies based on descriptive details in-dialogue. Table \ref{fig:trainingtasks} displays an example of a tag-to-movie mapping.

\subsection{MovieLens Reviews Task}

The final training task, referred to as the MovieLens Review task, uses a joint dataset created by \citet{penha2020what} to incorporate additional movie description and opinion data. The training examples for this task are generated from the reviews portion of \citet{penha2020what}'s search dataset, which contains the IMDB user reviews associated with each movie in the MovieLens database. These reviews contain movie attribute data written in the kind of casual, natural dialogue style found in the ReDial dataset, so they aid the model's natural text generation and descriptive capabilities. As shown in Table \ref{fig:trainingtasks}, these reviews are processed into examples where the model is asked to predict the next sentence of a review given a movie title and the truncated review \footnote{Because movie titles and title fragments in the MovieLens Reviews dataset are not delimited, the MovieLens Reviews training task does not use '@' signs to separate movie titles.}.

\subsection{Multitask Training}

The T5 module supports multitask training, where examples from training dataset are loaded through their own preprocessing steps (in our case only lowercasing). We opt to finetune the base size (220 million parameters) with a learning rate of 0.003 for 40,000 steps and batch size 128 \footnote{The T5 was finetuned with using the public T5 codebase: https://github.com/google-research/text-to-text-transfer-transformer}. Texts longer than the maximum sequence length, i.e., 512 for inputs and 128 for targets are truncated. We train variants with different combinations of the four training tasks in order to isolate their effects. Examples from each task were loaded equally often. As suggested by~\citet{raffel2019exploring}, we prepend the inputs to each task with a task label: ``redial conversation:", ``movielens sequence:", ``movielens tags:", or ``movielens review:". From this point, the name T5 will be used to refer to the out-of-the-box pretrained T5 model and the name T5-CR will be used to refer to our custom T5 model with all four finetuning tasks.

\section{Baseline Evaluations}

In order to determine whether our end-to-end approach can perform competitively on dialogue and recommendation, we compare our performance using BLEU score and Recall. These metrics are both run on the evaluation set provided with the ReDial dataset. The BLEU score acts as a measure of dialogue quality, by measuring the similarity of the model and human responses. The Recall metric is calculated by comparing the movies mentioned by the model in the evaluation dialogues to the set of movies mentioned by the human recommender in the ReDial dialogues evaluation set.  \textit{End-to-End Recall} refers to the Recall@1 caluclated in the dialogue task, while \textit{Rec Module Recall} is calculated on the same movies run through only the isolated recommendation module if one exists. These two metrics were selected as they are standards which have been run on many of the previous works in the area. We calculated baselines for the ReDial and KBRD models, which did not have reported BLEU or Recall scores, and sources our other baselines from \citeauthor{wang2021finetuning}. In Table \ref{fig:metricsresults} we compare the most relevant models: (1) \textbf{ReDial} \cite{li2018towards} an HRED CR system, (2) \textbf{KBRD} \cite{chen2019towards} which uses a transformer for dialogue and a knowledge graph for recommendation, (3) \textbf{KGSF} \cite{zhou2020improving} which uses fuses word and entity knowlege using a knowlege graph and transformer, (4) \textbf{GPT-2} a pretrained transformer, (5) \textbf{RID} \cite{wang2021finetuning} which uses a pretrained language model with a relational graph convolutional network (RGCN).

\begin{table*}[tb]\centering 
% \resizebox{\columnwidth}{!}{
\begin{tabular}{llrrr} 
{\bf Model Name} & {\bf Model Type} & {\bf BLEU} & {\bf Rec Module Recall} & {\bf End-to-End Recall}\\ \hline \hline
{\bf ReDial} & Autoencoder + LSTM  & 8.38 & 2.30 & 0.70\\ \hline
{\bf KBRD} & Knowledge Graph  + Transformer & 11.00 & 3.00  & 0.80\\ \hline
{\bf KGSF} & Knowledge Graph Semantic Fusion & ----- & 3.90  & 0.90\\ \hline
{\bf GPT-2} & Pretrained Transformer & ----- & -----  & 1.40\\ \hline
{\bf RID} & RGCN  + Pretrained Transformer & {\bf 20.70} & -----  & 3.10\\ \hline
{\bf T5-CR} & Finetuned T5 (4 Tasks) & 15.39 & ----- & {\bf 6.93} \\
\end{tabular}
% }
\caption{BLEU and Recall@1 metric comparisons between T5-CR, our T5 variant finetuned on 4 tasks, and the previous approaches to conversational recommendations. All evaluation scores are calculated based on the model's performance on the ReDial validation dialogues. The baseline scores for the KGSF, GPT-2, and RID models along with the End-to-End Recall scores for the KBRD and ReDial models were taken from \cite{wang2021finetuning}.}
\label{fig:metricsresults} 
\end{table*}

\subsection{BLEU}
    BLEU score is a standard metric used in machine translation and text generation tasks which quantifies how similar a generated phrase is to the expected target phrase \cite{10.3115/1073083.1073135}. We postprocess our ReDial predictions to replace movie titles with a \texttt{"\_\_unk\_\_"} token before calculating the metric. This ensures that our BLEU score only captures information on the closeness of the dialogue to our target, and isn't influenced by correct/incorrect movie titles and recommendations. Our T5-CR model, trained on all four training tasks was able to outperform the KBRD and ReDial models approaches, achieving a BLEU score of 15.39. The increase in BLEU score is likely a result of the introduction of movie description and attribute data through the multitask training setup as well as general increased fluency of large pre-trained language models such as the T5. The RID model trained by \citeauthor{wang2021finetuning} was also built using a pretrained tranformer and performed the best, with a score of 20.70.
    
\subsection{Recall}
    In order to evaluate the quality of the recommendations we calculate End-to-End Recall as the percent of movies generated by the model in-dialogue which correspond to one of the known recommendations given by the human recommender in the corresponding redial dialogue. Rec Module Recall refers to the Recall@1 score when only the isolated recommendation modules are used. In previous efforts such as the ReDial, KBRD, and KGSF models, the End-to-End Recall scores were significantly lower than the Rec Module Recall scores, suggesting that the models were less likely to apply their recommendation knowledge accurately in-dialogue compared to as a separated recommendation task. This highlights the advantage of the end-to-end approach, as the unified structures ensures the model can generate high quality dialogue and recommendations simultaneously. The multitask T5-CR model achieved a Recall score of 6.93, which outperforms all baseline models. The increase in Recall is likely due to the end-to-end structure of the model allowing it to use dialogue features to retrieve better recommendations, as well as the movie relationship and attribute training tasks allowing for more accurate analysis of user preferences.

\section{Probe Studies}
Although the BLEU and Recall scores on the ReDial Evaluation Dataset prove that an end-to-end model can outperform multi-component approaches, the scores do not give us insight on the extent to which our multitask training setup benefited the model's ability to generate dialogue and recommendations. Also, the ReDial evaluation slice covers a small selection of movies and dialogue interactions. In order to determine the contribution of each of the training tasks, as well as any measurable advantages of cross-task transfer within the same T5 architecture, we present four probe studies in the style of \citet{penha2020what}. Each probe tests a specific dialogue interaction by measuring the T5-CR's ability to distinguish between relevant and unrelated information conditioned on different types of sample dialogues. In order to filter out misspellings, alternate titles, and rare movies which the model has little information on, the probes are generated using the set of movies which occur over 30 times in the ML Sequences dataset. With probe examples generated from this set of around 5,000 movies, we are able to run evaluations on a much larger range of data than the limited ReDial evaluation set. These probes are designed to measure the model's ability to apply the information gained through its multitask training in a dialogue setting, therefore all probe data is evaluated through the ReDial Dialogue tasks. 

\begin{table*}[tb]\centering 
% \resizebox{\columnwidth}{!}{
\begin{tabular}{llllll}
& {\bf T5 Finetuning Tasks} & {\bf Rec Probe} & {\bf Attr Probe} & {\bf Combo Probe} & {\bf Desc Probe} \\ \hline \hline
 & None (\bf{T5}) & 0.5493 & 0.4908 & 0.5597 & 0.5936 \\ \hline
 & ReDial & 0.4716 & 0.5046 & 0.5731 & 0.7097 \\ \hline
& ReDial + & & & & \\
 & ML Sequences & 0.6359 & 0.5869 & 0.7367 & 0.7307 \\ \hline
& ReDial + & & & & \\
 & ML Tags & 0.5670 & {\bf 0.7826} & 0.8016 & 0.7133 \\ \hline
& ReDial + & & & & \\
 & ML Reviews & 0.4771 & 0.5091 & 0.5833 & 0.7763 \\ \hline
 & All (\bf{T5-CR}) & {\bf 0.6599} & 0.7678 & {\bf 0.8418} & {\bf 0.7928} \\
\end{tabular}
% }
\caption{Comparison of probe scores across T5 models with different finetuning tasks.}
\label{fig:proberesults} 
\end{table*}

\subsection{Recommendation Probe}
The recommendation probe measures the model's ability to distinguish a related movie from a popular movie chosen at random. In order to quantify related movies based on cooccurrence in the ML Sequences dataset, we rank movies based on $PMI^2$ \cite{role2011handling}, a variation on pointwise mutual information ($PMI$). $PMI^2$ is a commonly used variation on $PMI$ which reduces $PMI$'s known bias toward rare and infrequent items \cite{role2011handling}. For each of the top ten related movies we sample a random popular movie from the top $10\%$ of movies (ranked by frequency in the ML Sequences dataset). For each of the ten $(related_i, \  popular_i)$ pairs generated for each movie, we create a probe by swapping in the movies to a generic piece of dialogue, as seen in Table \ref{fig:probeexamples}. The probe score is calculated as the percent of probes where the models log-likelihood score, $L(\theta)$, of  the target containing the related movie was higher than that of the random popular movie. Note that different phrasings and dialogue formats were tested with little effect on the probe results.

As shown in Figure \ref{fig:proberesults}, the introduction of the ML Sequences task improved the model's ability to differentiate between related and random movies, reflected by a $30\%$ increase in the recommendation probe scores between the ReDial-only model and the ReDial + ML Sequences model. This increase demonstrates that the patterns in the movie sequences fed in the ML Sequences tasks can be generalized and applied within the dialogue tasks. Interestingly, the ReDial + ML Tags model also outperformed the ReDial-only model, with an increase of $23\%$ in recommendation probe scores over the ReDial-only model.

This increase demonstrates an advantage of the end-to-end format: data incorporated to help the understanding of descriptive words in the dialogues also boosted performance on movie-to-movie recommendation, despite the additional data not directly specifying any movie relationships. Because recommendation and dialogue are handled in the same model, it can leverage patterns in seemingly unrelated data. Here, the model is likely using the overlap of tags associated with different movies to help determine whether they are related. In the combined model, where all four training tasks were included, the model performed the best ($+37\%$ over ReDial-only), a
score which demonstrates the viability of multitask transformers to incorporate many different data sources and tasks without losing performance.

Overall the performance of the recommendation probe represents a transfer of movie-relationship knowledge between training tasks, but this transfer is not perfect. While the probes (fed into the ReDial Dialogues task) achieved a score of $.6599$ in the Combined model, the same pairs fed into the ML Sequences task without any dialogue achieved a score of $.7711$. This increase indicates either an incomplete transfer of knowledge from the MovieLens Sequences task to the ReDial Dialogues task, or a bias from the the movie recommendation data already present in the ReDial Conversations. Similarly, the T5's performance on the movie recommendation probes is lower than that of a purely Matrix Factorization model, which achieved a score of $.8096$ on the movie pairs. 

\subsection{Attributes Probe}

The attributes probe measures the model's ability to use details and descriptive words appearing in-dialogue to retrieve relevant movies. As shown in Table \ref{fig:probeexamples}, a probe is generated for each movie-tag association in the MovieLens Tags dataset, with a random popular movie used as the negative. Because many of the most popular tags (such as "action" or "excellent") might apply to a large portion of the popular movies, we filter the negative to ensure it isn't associated with the given tag.

The attribute probe scores also demonstrated the effectiveness of multitask learning, with the introduction of the ML Tags task leading to a $52\%$ increase in performance over the ReDial-only model. This probe directly shows one of the advantages of end-to-end learning. Because dialogue analysis and recommendation generation occurs in the same model, the descriptive attributes mentioned in the input dialogue (or movie "tags") can help the model retrieve a movie relevant to that attribute, even when no movie titles are mentioned in the input. While the Combined model didn't out-perform the RD Tags model, it did perform consistently, with an accuracy of $.7689$ over the probe set. 

\begin{table*}[t]\centering \resizebox{\textwidth}{!}{
\begin{tabular}{p{0.1\textwidth}p{0.25\textwidth}p{0.23\textwidth}p{0.22\textwidth}p{0.2\textwidth}} 
& \bf{Recommendation Probe} & \bf{Attribute Probe} & \bf{Combination Probe} & \bf{Description Probe} \\ \hline \hline
\begin{tabular}[c]{@{}l@{}}\\ \textbf{Input 1}\\(Related)\end{tabular} 
& [User] Can you recommend me a movie like \movie{@ Zootopia (2016)} @
& [User] Can you recommend me a vampire movie? 
& [User] Can you recommend me a science fiction movie like \movie{@ Looper (2012) @}?
&  [User] What is your opinion on \movie{@ Ringing Bell (1978) @}?  \\ 
\begin{tabular}[c]{@{}l@{}}\\[13pt] \textbf{Input 2}\\(Rand.\\ Popular)\end{tabular} & & & & [User] What is your opinion on \movie{@ Robin Hood: Men in Tights (1993) @}? \\ \hline
\begin{tabular}[c]{@{}l@{}}\\ \textbf{Target 1}\\ (Related)\end{tabular}  & Sure, have you seen \movie{@ Inside Out (2015) @}?
& Sure, have you seen \movie{@ Interview with the Vampire: the Vampire Chronicles (1994) @}? &  Sure, have you seen \movie{@ Edge of Tomorrow(2014) @}?  & Watching this several times as a child was quite the… \\ 

\begin{tabular}[c]{@{}l@{}}\\[13pt]\textbf{Target 2}\\(Rand.\\ Popular)\end{tabular}
 & Sure, have you seen \movie{@ I Am Sam (2001) @}?  & Sure, have you seen \movie{@ Sicko (2007) @}? &  Sure, have you seen \movie{@ Zoolander (2001) @}? & \\ \hline
\textbf{Data}  & ML Sequences &  ML Tags &  ML Sequences + Tags & ML Reviews \\ \hline
\textbf{Metric}  
& \begin{tabular}[c]{@{}l@{}}$L(T_1 \ | \ I_1) >$ $ L(T_2 \ | \ I_1)$\end{tabular}  
& \begin{tabular}[c]{@{}l@{}}$L(T_1 \ | \ I_1) >$ $ L(T_2 \ | \ I_1)$\end{tabular} 
& \begin{tabular}[c]{@{}l@{}}$L(T_1 \ | \ I_1) >$ $ L(T_2 \ | \ I_1)$\end{tabular}
& \begin{tabular}[c]{@{}l@{}}$L(T_1 \ | \ I_1) >$ $ L(T_1 \ | \ I_2)$\end{tabular} \\
% & $L(T_1 \ | \ I_1) > L(T_1 \ | \ I_2)$ \\ \\

\end{tabular}
}
\caption{Comparison of the four probe sets, which determine whether the model can correctly rank related entities as more likely than random negatives.}
\label{fig:probeexamples} 
\end{table*}

\subsection{Combination Probe}

The combination probe measures the multitask capabilities of the model, determining whether attribute and movie entity data can be used simultaneously to generate a relevant response. As shown in Table \ref{fig:probeexamples}, a probe is generated for each shared tag among each of a movies top 10 most related movies. As in the attribute probe, we filter out the popular negative to ensure it does not match the given tag.

The combination probe extends the findings of the previous two probes: not only can the model use mentioned movies or movie attributes to influence its recommendations, it can do both at the same time. Whereas a multi-component approach to the problem would base its recommendation solely on the previously mentioned movies or the attributes mentioned in-dialogue, an end-to-end approach uses these pieces of information together. The Combined model was able to differentiate $84.18\%$ of the probe pairs when given a movie and a tag in the input dialogue, an improvement over its performance on either the recommendation or attribute probes. This improvement demonstrates that when using both types of information together, the model can more accurately recommend a related movie.

\subsection{Movie Description Probe}

The previous three probes test whether the model can retrieve a relevant movie title when conditioned on a dialogue. The movie description probe tests the reverse direction: can the model retrieve a piece of relevant or descriptive dialogue when conditioned on a certain movie title. To do this, we measure the likelihood of a given review snippet taken from the first four sentences of a review in the ML Reviews dataset. In previous probes, we have ranked two different targets based on likelihood, but because review snippets differ greatly in length, phrasing, style of language, and other factors which can influence likelihood, we opt to keep the target the same and compare the likelihood of a given review snippet when conditioned on a related/unrelated movie. As shown in Table \ref{fig:probeexamples}, for a related input $I_1$, a random popular input $I_2$, and a review snippet $T$ we compare the log likelihood scores and measure how often $L(T\  |\  I_1) > L(T\  |\  I_2)$. 

The description probe demonstrates that in an end-to-end model, mentioning a movie can prompt the model to retrieve relevant dialogue. This functionality wouldn't be in traditional multi-component approaches where mentioned movies are processed separately from dialogue. The ML Reviews training task led to a $9.38\%$ increase over the ReDial-only model, while the combined model was able to achieve a score of $0.7929$, an $11.72\%$ increase over the ReDial-only model.  

\section{Conclusion}

In this paper, we presented a multitask approach to end-to-end conversational recommendations. In direct comparison to two previously published models in the domain, our T5-based architecture outperformed the baselines in both its quality of dialogue and recommendation. When probed on recommendation, attribute knowledge, and description, our model demonstrates that dialogues and recommendations can be mutually improved by sharing a model architecture. Specifically, the probes prove that the model is able to use dialogue features to inform its recommendations and movie mentions to influence its dialogue generation. These findings support a general trend in current natural language processing landscape, where large pretrained transformer models are rapidly becoming the state-of-the-art in many domains. In fact, our research has implication on the broader area of multitask models, highlighting how a limited dataset (such as ReDial) can be injected with information from several auxiliary datasets, regardless of format. In the future, this effect might shift the focus from training and combining optimized components for each functionality of a system, to simply incorporating all desired information as different tasks in a pretrained multitask model. 
% References and End of Paper
% These lines must be placed at the end of your paper
\bibliography{refs}
\bibliographystyle{acl_natbib}

\clearpage

\appendix

\section{Appendix - Redial Dataset Sample}
An example of the beginning of a ReDial con-versation, randomly selected from the ReDial dataset is shown at ~\ref{fig:redialconv}.

\label{sec:appendix}
 \begin{figure}[!th]
 \begin{mdframed}[backgroundcolor=black!8,rightline=false,leftline=false]
    \begin{small}
    {\bf \texttt{REDIAL CONVERSATION:}} \\ \\
    \texttt{{\bf \texttt{Sender:}} I'm in the mood to watch a romantic comedy. What do you suggest?} \\ \\ 
    \texttt{{\bf \texttt{Responder:}} \movie{@ 50 First Dates (2004) @} Have you seen that one?} \\ \\
    \texttt{{\bf \texttt{Sender:}} Oh, I've seen that one. I really like...}
    
    \end{small}
    \end{mdframed}
    \caption{Beginning of a randomly selected example from the ReDial dataset.}
    \label{fig:redialconv}
\end{figure}
\end{document}